\documentclass[letterpaper]{article} 
\usepackage{aaai2026}  
\usepackage{times}  
\usepackage{helvet}  
\usepackage{courier}  
\usepackage[hyphens]{url}  
\usepackage{graphicx} 
\usepackage{natbib}  
\usepackage{caption} 
\usepackage{algorithm}
\usepackage{algorithmic}

\usepackage{xcolor}
\usepackage{soul}
\usepackage{amsmath}
\usepackage{multirow}
\usepackage{booktabs}
\usepackage{newfloat}
\usepackage{listings}
\DeclareCaptionStyle{ruled}{labelfont=normalfont,labelsep=colon,strut=off} 
\floatstyle{ruled}
\newfloat{listing}{tb}{lst}{}
\floatname{listing}{Listing}
\title{Physics-Informed Autonomous LLM Agents for Explainable Power Electronics Modulation Design}
\author{
    Junhua Liu\textsuperscript{\rm 1},
    Fanfan Lin\textsuperscript{\rm 2},
    Xinze Li\textsuperscript{\rm 3}\thanks{Corresponding author.},
    Shuai Zhao\textsuperscript{\rm 4},
    Kwan Hui Lim\textsuperscript{\rm 1,5}
}
\affiliations{
    \textsuperscript{\rm 1}Forth AI, Singapore\\
    \textsuperscript{\rm 2}Zhejiang University - University of Illinois Urbana-Champaign Institute, China\\
    \textsuperscript{\rm 3}University of Arkansas, USA\\
    \textsuperscript{\rm 4}Aalborg University, Denmark\\
    \textsuperscript{\rm 5}Singapore University of Technology and Design, Singapore\\

    j@forth.ai, fanfanlin@intl.zju.edu.cn, xinzel@uark.edu, szh@energy.aau.dk, kwanhui@acm.org
}

\usepackage{bibentry}

\begin{document}

\maketitle

\begin{abstract}
LLM-based autonomous agents have recently shown strong capabilities in solving complex industrial design tasks. However, in domains aiming for carbon neutrality and high-performance renewable energy systems, current AI-assisted design automation methods face critical challenges in explainability, scalability, and practical usability. To address these limitations, we introduce PHIA (Physics-Informed Autonomous Agent), an LLM-driven system that automates modulation design for power converters in Power Electronics Systems with minimal human intervention. In contrast to traditional pipeline-based methods, PHIA incorporates an LLM-based planning module that interactively acquires and verifies design requirements via a user-friendly chat interface. This planner collaborates with physics-informed simulation and optimization components to autonomously generate and iteratively refine modulation designs. The interactive interface also supports interpretability by providing textual explanations and visual outputs throughout the design process. Experimental results show that PHIA reduces standard mean absolute error by 63.2\% compared to the second-best benchmark and accelerates the overall design process by over 33 times. A user study involving 20 domain experts further confirms PHIA’s superior design efficiency and usability, highlighting its potential to transform industrial design workflows in power electronics.
\end{abstract}


\renewcommand{\arraystretch}{1.05}

\section{Introduction}

As global temperatures rise due to climate change, transitioning power generation systems toward carbon neutrality is increasingly urgent. A key component of this transition is the integration of renewable energy sources (RES) into the power grid. Power Electronics Systems (PES) are critical to this effort for three primary reasons~\cite{lin2022dual}.
Firstly, power converters within PES transform the direct current (DC) output from solar panels or wind turbines to alternating current (AC), facilitating the seamless integration of RES into the AC grid. Secondly, PES manages power during peak and off-peak periods, guiding the charge and discharge processes of energy storage units. Lastly, it upholds grid stability through functions like voltage regulation and reactive power compensation. Achieving the desired outcomes necessitates meticulous modulation of power converters in PES, which involves controlling their switching to manage output voltage or current.

The escalating adoption of RES and the burgeoning scale of multi-resource power systems complicate the modulation design for power converters within PES. The diversity of energy sources linked to the power grid intensifies system complexity. As more converter subsystems interlink, the modulation design problem's dimensionality surges. Consequently, identifying an optimized design solution through manual computation becomes infeasible given this complexity. As cities worldwide embrace multi-resource power systems, modulation designs must be customized, catering to particular geographies and application objectives. Oftentimes, these objectives can conflict, further complicating matters.

The recent advancements in artificial intelligence (AI) and Large Language Models (LLMs) offer a promising solution to these modulation design challenges through automation. Emerging research has introduced AI-centric modulation design techniques. For instance, recent works explore the use of an LLM with in-context learning for modulation design ~\cite{lin2024pe}. Extreme gradient boosting algorithm (XGBoost) was utilised to construct the surrogate model for the triple phase shift modulation strategy for the dual active bridge (DAB) converter using training data sourced from simulation tools or hardware prototypes~\cite{lin_ai-based_2023}. Subsequently, a differential evolution algorithm collaborates iteratively with XGBoost until optimal modulation parameters are identified. Likewise, the Q-learning algorithm, a conventional reinforcement learning technique, can be trained offline to derive optimized modulation parameters \cite{9138774}. 
While innovative, these methods suffer from several drawbacks:
\begin{enumerate}
\item The training of these AI models is often data-intensive, with significant demands for an extended and tedious data collection through extensive simulations or hardware experiments. 
\item For complex tasks, training or finetuning large models are also computationally intensive and consume a large amount of energy to train and serve.
\item Deploying AI models as unexplainable black boxes severely restricts their industrial adoption.
\item The current techniques focuses on specific modulation strategies or pre-established design goals, which limits the scalability of such automation.
\item Existing methods require extensive human involvement in the whole design process which makes the design process inefficient.
\end{enumerate}

In recent years, large language models, such as GPT-4~\cite{openai2023gpt4}, Palm~\cite{anil2023palm} and LLaMa~\cite{touvron2023llama} demonstrates superior performance in natural language understanding and generation. Furthermore, recent works introduce LLM-based autonomous agents which extend LLM's capacity from simply performing reasoning and generating content to actions and control~\cite{wang2024survey}. 

In our work, we investigate the potential of using LLM-based autonomous agents for downstream task automation in Power Electronics Systems. Specifically, we introduce \textbf{PHIA}, a Physics-Informed Autonomous Agent for power converter modulation design, enabling users to produce high quality modulation designs with minimal human supervision. 

PHIA first uses a LLM-based planner to process user requirements and generate a set of design specifications. Subsequently, the agent invokes a set of system design tools that consist of physics-informed surrogate models and optimization algorithms. In the design process, the tools iteratively derives optimal modulation parameters tailored to users' design specifications. Finally, the optimal design parameters and performance metrics will be returned to and visualised for the users to achieve an explainable design process. 

The contributions of this paper are summarized as follows:
\begin{itemize}
\item We propose PHIA, a physics-informed autonomous agent that streamlines the power converter modulation design process. Comprehensive experiments shows strong performance of PHIA, statistically outperforming the best baseline model with 63.2\% lower error for low-data scenario, and 23.7\% lower error for high-data scenario.
\item We propose a hierarchical physics-informed surrogate model as a design automation tool for power converter modulation, which outperforms benchmarks in accurately predicting power converter performance even in extreme scenarios where the data is sparse.
\item Our user study with 20 experts demonstrates that the design time with PHIA is over 33 times faster than conventional methods, showing superior efficiency in empirical practice with minimal human supervision.
\end{itemize}
\begin{figure}[t]
    \begin{center}
        \includegraphics[width=1\linewidth]{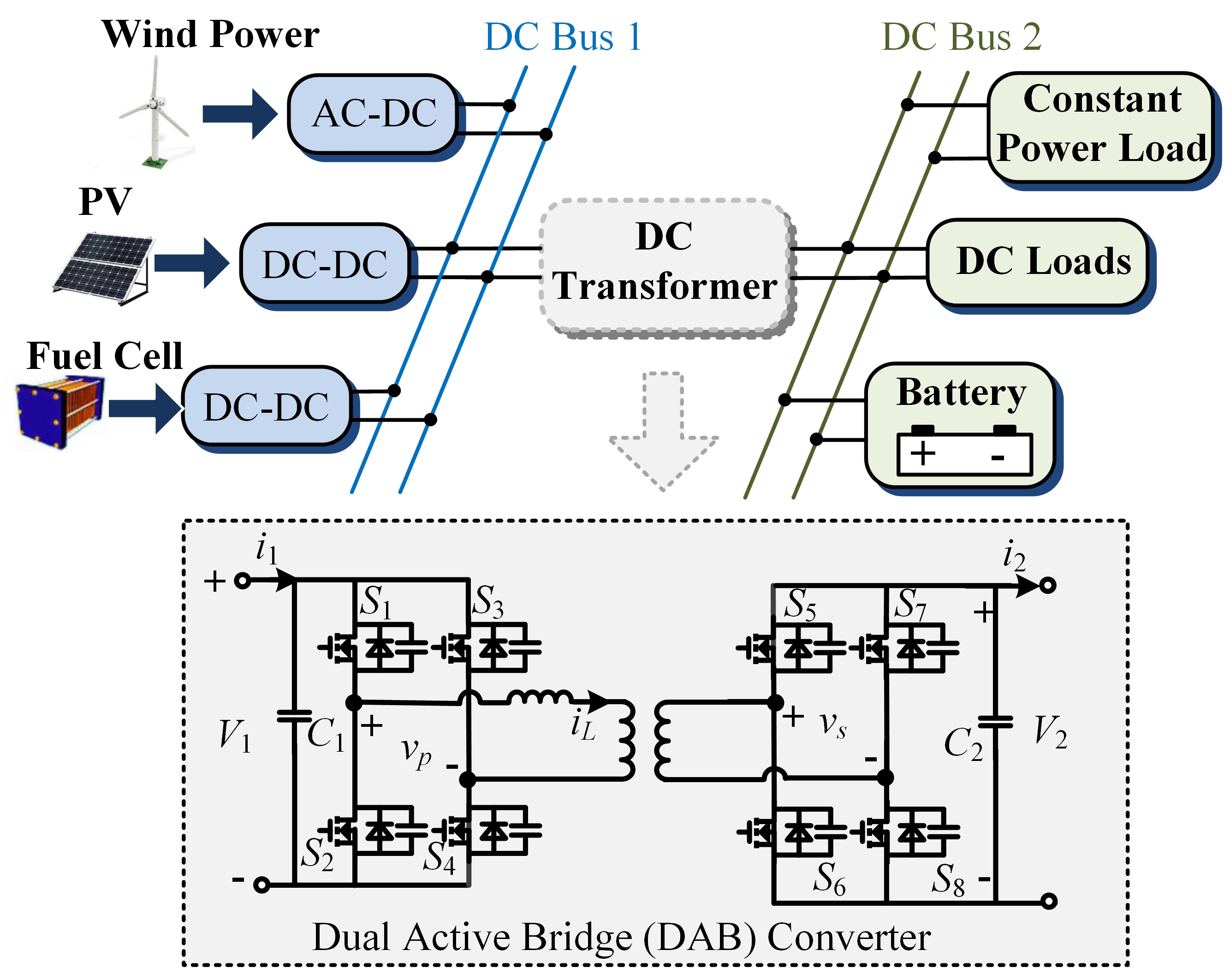}
    \end{center}
    \setlength{\abovecaptionskip}{-0.03cm}
    \caption{An example of power converter application: the DAB (Dual Active Bridge) converter serves as a DC transformer of the DC-DC power grid, offering galvanic isolation and regulating power and voltage between DC buses. Modulating its switches directly affects the system operating performance, including power transfer efficiency, voltage regulation, and stability of the interconnected buses.}
    \label{fig:1}
\end{figure}
\begin{figure*}[t]
    \begin{center}
        \includegraphics[width=1.0\linewidth]{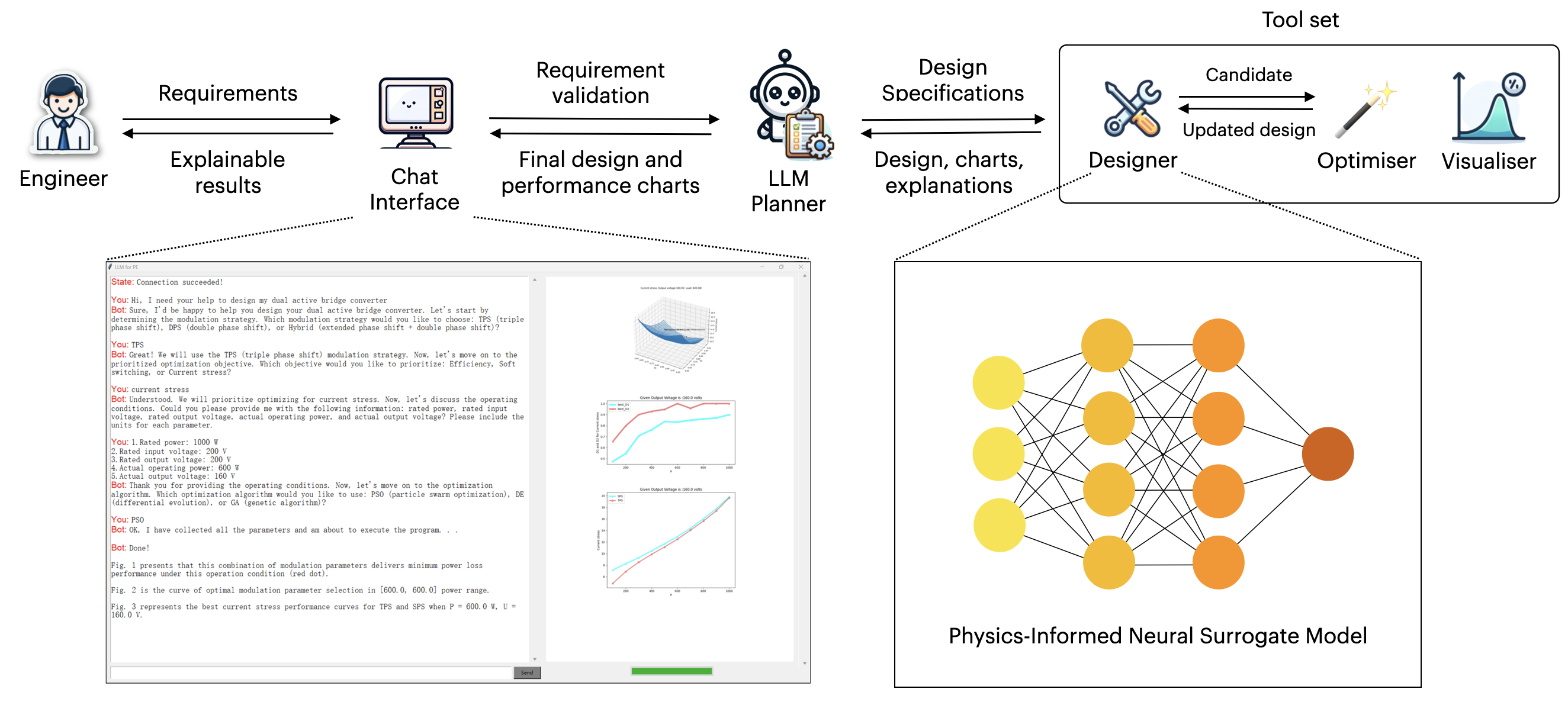}
    \end{center}
    \setlength{\abovecaptionskip}{-0.03cm}
    \caption{System architecture of PHIA: an engineer provides design requirements to PHIA via a chat interface connecting to its planner. Once the full requirements are determined, the planner coordinates and invokes tools from the tool set to iteratively generate the modulation design without human supervision. After the design is done, the planner displays the final results and explainable process on the chat interface.}
    \label{fig:2}
\end{figure*}

\section{Problem Statements}

\subsection{Power Converter Phase Shift Modulation}
Within the renewable energy systems sector, power converters need to work in coordination with other components to ensure optimal functionality. Figure~\ref{fig:1} depicts a classic application of a particular power converter, the dual active bridge (DAB) converter. This converter functions as a DC transformer, connecting various DC buses to facilitate the utilization or storage of power produced by RES \cite{9356241}. The phase shift modulation technique is employed to regulate the current flow and guarantee efficient power transmission. This is done by adjusting the timing of different power switch sets, represented as $S_1$ to $S_8$ in Figure~\ref{fig:1}. Table~\ref{table:PhaseShift} details the distinct phase shift modulation strategies associated with the DAB converter, highlighting their respective degrees of freedom (DoF) and modifiable parameters \cite{hou2019overview}. It is worth noting that the adjustable parameter for $<S_1, S_5>$ indicates the phase difference between $S_1$ and $S_5$.

\begin{table}[t]
\begin{center}
\scalebox{0.77}{
\begin{tabular}{ccccc}
\toprule
\textbf{\begin{tabular}[c]{@{}c@{}}Phase shift \\ Strategies\end{tabular}} & No. of DoF & \textbf{$<S_1, S_3>$} & \textbf{$<S_1, S_5>$} & \textbf{$<S_5, S_7>$} \\ \midrule
\begin{tabular}[c]{@{}c@{}}Single phase \\ shift\end{tabular} & 1 & 0 & $D_o$ & 0 \\ \midrule
\begin{tabular}[c]{@{}c@{}}Double phase \\ shift\end{tabular} & 2 & $D_i$ & $D_o$ & $D_i$ \\ \midrule
\begin{tabular}[c]{@{}c@{}}Extended phase \\ shift 1\end{tabular} & 2 & $D_i$ & $D_o$ & 0 \\ \midrule
\begin{tabular}[c]{@{}c@{}}Extended phase \\ shift 2\end{tabular} & 2 & 0 & $D_o$ & $D_i$ \\ \midrule
\begin{tabular}[c]{@{}c@{}}Triple phase \\ shift\end{tabular} & 3 & $D_1$ & $D_o$ & $D_2$ \\ \midrule
Hybrid & 2 & $D_i$ & $D_o$ & 0 \\  
phase  & 2 & 0 & $D_o$ & $D_i$ \\ 
shift & 2 & $D_i$ & $D_o$ & $D_i$ \\ \bottomrule
\end{tabular}}
\end{center}
\caption{Phase shift modulation strategies with the number of degrees of freedom (DoF) from 1 to 3. {$<S_x, S_y>$} defines the phase shift between $S_x$ and $S_y$.}
\label{table:PhaseShift}
\vspace{-.5cm}
\end{table}

For designing the parameters of the modulation strategy, users are required to define three essential pieces of specification consistent with the application prerequisites, which are:

\begin{enumerate}
    \item Expected operating conditions for the converter: This involves specifying the rated power, rated input voltage (current) and rated output voltage (current), as well as their actual operating values.  

    \item Selected modulation strategy: The modulation strategy selection should factor in both the acceptable level of control complexity and the feasibility of achieving the desired performance for the specific strategy. 

    \item Performance objectives: Various application scenarios have different priorities for performance objectives, which directly affect the modulation design. Common objectives in practical applications encompass power efficiency, power density, zero voltage switching, zero current switching, current stress, dynamic response, electromagnetic interference, stability, and compatibility with other subsystems, etc.  
\end{enumerate}

\subsection{Problem Statements}
Given a set of operating conditions of the power converter (rated power $P_r$, actual power $P_a$, rated input voltage $V_{1r}$, rated output voltage $V_{2r}$, actual operating output voltage $V_{2a}$), chosen modulation strategy $S$ and prioritized performance objectives $O$, the aim is to design modulation parameters for the chosen modulation strategy $S$ to achieve optimal performance for $O$. This design system is expected to have the following features:  
\begin{itemize}
    \item Automatic design outcome generation, eliminating the need for users to perform complex analysis or computations.  
    \item Precise design outcomes that translate into exceptional operational performance in real-world scenarios. 
    \item Good usability and explainability to accelerate engineering design and scientific discovery.
    \item Simple setup for the design system, minimizing the demand for extensive data resources.  
\end{itemize}
\begin{figure*}[t]
    \begin{center}
        \includegraphics[width=1.0\linewidth]{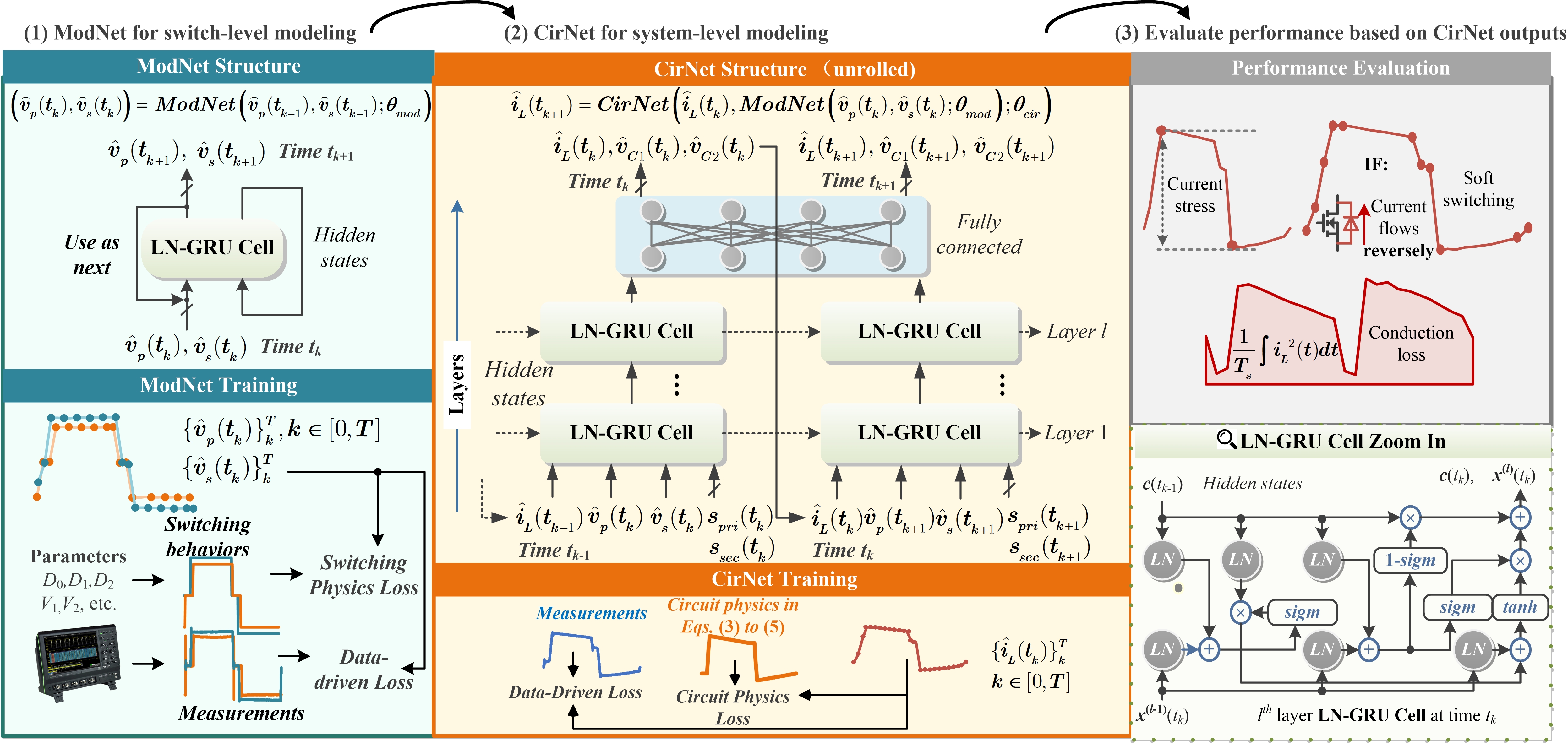}
    \end{center}
    \setlength{\abovecaptionskip}{-0.03cm}
    \caption{The proposed surrogate model consists of two physics-informed neural networks, namely, \textbf{ModNet} for switch-level modeling to learn the switching behaviors, and \textbf{CirNet} for system-level modeling to learn the circuit physics. The hierarchical structure enhances the overall accuracy of the power converter's modeling of the complex behaviors of the switches.}
    \label{fig:3}
\end{figure*}

\section{Related work}

\subsection{LLM-based Autonomous Agents}
The evolution of Large Language Models (LLMs) has unlocked immense potential for automating various processes in engineering and natural sciences through simple, plain-language queries. Numerous studies have explored LLM capability in reasoning and formal system applications, such as programming co-pilots~\cite{ross2023programmer,mcnutt2023design}, multi-lingual tasks~\cite{xu2024face4rag,liu2024lara}, and mathematical problem-solving \cite{shakarian_independent_2023,shi2024math}. 

To address more complex industrial tasks, recent work proposed the use of LLM-based autonomous agents (LLM-agents), which extends the capability of LLMs~\cite{wang2024survey}. LLMs are deployed as the planning backbone to understand user request and invoke suitable tools to perform actions~\cite{yao2024llm}. Besides the planning module, LLM-agents also include other components, such as memory and tools. With advanced techniques such as Retrieval Augmented Generation (RAG)~\cite{liu2024lara,xu2024face4rag}, LLM-agents become powerful tools for solving complex tasks with little to no human instruction. 

Recent works reported outstanding performance of LLM-agents in industrial applications. For instance, Cao and Lee harnessed LLMs and Phase-Step prompts for cross-domain robot task generation \cite{cao_robot_2023}. Li et al. introduced an agentic framework to for strategic and interactive decision-making~\cite{li2024stride}. LLM-based agents are found effective in other domain-specific tasks, such as spatial-temporal strategic planning~\cite{li2024urbangpt, liu2024spatial, xiang2024performative}, e-commerce customer supports~\cite{liu2024lara,liu2026balancing,liu2025intents} and financial trading~\cite{zhang2024multimodal,koa2024learning,zhang2024ai,gao2023stockformer}.

\subsection{Physics-Informed Neural Networks}
Conventional neural networks are commonly data-hungry and not explainable. By integrating physics principles into neural networks \cite{raissi_deep_2018}, studies on Physics-Informed Neural Networks (PINNs) show promising results in overcoming such challenges. 

PINN encompasses various approaches to incorporate physics-informed (PI) components into the deep learning pipeline, such as loss functions, parameter initializations and the neural network architectures \cite{huang_applications_2023}. As a result, the neural networks are able to learn latent features that are governed by physical laws. 

Recent studies also enhance PINN's convergence, stability, and accuracy, yielding improvements \cite{kang2023pixel,yang2023dmis,wandel2022spline}. PINN has found success in a wide range of applications. Some examples include predicting traffic conditions and modeling traffic flow \cite{shi_physics-informed_2021,ji2022stden}, determining circuit parameters for power converter health monitoring \cite{zhao2022parameter}, computing power flow in electrical systems \cite{hu_physics-guided_2021}, modeling temperature dynamics in lakes \cite{jia_physics-guided_2020}, managing real-time reservoir gas production \cite{mudunuru_physics-informed_nodate}, designing quantum circuits~\cite{liu2021hybrid,guo2019general}, and simulating turbulent flows in urban environments \cite{xiao_reduced_2019}. These applications highlight PINNs' promising performance across many domains by infusing physics knowledge into deep learning.

\section{Methodology}


\subsection{Agentic AI System}
Figure~\ref{fig:2} presents the architecture of the proposed PHIA agentic AI system. The front-end consists of a chat interface and API interaction with GPT-4 as reasoning engine, whereas the back-end consists of the agent's tooling, i.e., an optimization algorithm and the PINN-based surrogate model. In terms of workflow, PHIA first collects design specifications from the user and passes to an LLM-based reasoning engine (GPT-4) for planning. Subsequently, the agent invokes the back-end tooling, where the surrogate model produces the performance metrics and passes to the optimization algorithm searches for the optimal modulation parameters. The back-end will respond with a set of optimal design parameters to the front-end to display and visualise as charts for explainability and readability.

\subsection{Physics-Informed Power Electronics Modeling}

Power electronics systems are highly nonlinear induced by the complex switching behaviors of semiconductor devices and the coupling of energy storage components. The generic representation of a power electronics system formulated by the nonlinear time-variant state-space equations is given in Eq.~\ref{algo:dudt}, where $\theta$ are the circuit parameters, $u(t)$ and $x(t)$ are the state variables and the input variables, $g(\cdot)$ are general nonlinear functions governing state transition behaviors, and $h(\cdot)$ denotes the input physical laws.

Overall, the proposed surrogate model consists of two hierarchical PINNs, one ModNet for switch-level modeling to learn the switching behaviors and another one CirNet for the system-level modeling to learn the circuit physics. ModNet is trained to learn the intermediate waveforms $v_p$ and $v_s$ and CirNet infers key characteristic waveforms in the circuit including $i_L$, $v_{c1}$, $v_{c2}$, with which the operating performance such as the current stress, soft switching range, and efficiency can be evaluated. As shown in Figure~\ref{fig:1}, $v_p$ and $v_s$ denote the ac terminal voltages of the primary and secondary full bridges, respectively, and $i_L$, $v_{c1}$, and $v_{c2}$ represent the inductor current, input capacitor voltage, and output capacitor voltage, respectively. 

\begin{equation}
\frac{\partial u(t)}{\partial t} = g(x(t);\theta)u(t)+h(x(t);\theta)
\label{algo:dudt}
\end{equation}

\textbf{ModNet for Switch-level Modeling of Switching Behaviors}: In the modeling of semiconductor switches, the non-ideal equivalent circuit considers the parasitic inductances, capacitances, and resistances. The complicated interactions within circuit components result in nontrivial oscillations and overshoots, which will affect the overall operating performance of the power converter. To address this, the switching behaviors is proposed to be modeled by a physics-informed network, ModNet. ModNet is composed of several layers of gated recurrent units with layer normalization (LN-GRU). The temporal feature of LN-GRU accords well with the task of modulation waveform prediction. As shown in Figure~\ref{fig:3} Part (1) and Eq.~\ref{algo:ModNet}, ModNet infers $v_p(t_k)$ and $v_s(t_k)$ based on the information of the previous timestamp $t_{k-1}$ and the hidden states, and all the intermediate predictions within a switching cycle are stored for training. The training of ModNet utilizes two kinds of losses: (1) the loss based on physics information which helps with switching synchronization; (2) the loss based on experimental data captured by oscilloscope, which helps the model to capture oscillations and overshoots in practical operations.

\begin{equation}
\begin{aligned}
    (\hat{v}_p(t_k), \hat{v}_s(t_k)) = \operatorname{ModNet}( \hat{v}_p(t_{k-1}),\hat{v}_s(t_{k-1}); \theta_{mod})
\end{aligned}
\label{algo:ModNet}
\end{equation}

where $\hat{v}_p(t_k)$ and ${\hat{v}_s(t_k)}$ denote the ModNet predictions for ${v_p}$ and ${v_s}$ at time ${t_k}$ given the previous voltages $\hat{v}_p(t_{k-1})$ and ${\hat{v}_s(t_{k-1})}$. 

\begin{equation}
L \frac{\partial i_L(t)}{\partial t}=-R_L i_L(t)+v_p(t)-n v_s(t)
\label{algo:LDel}
\end{equation}

where $L$, $R_L$, and $n$ denote features of magnetic components: leakage inductance, equivalent inductor resistance, and turn ratio of transformer, respectively. $i_L(t)$ is the key waveform characterizing main circuit performances such as power transfer, efficiency, thermal behavior, soft switching, electromagnetic interference, etc.

\begin{equation}
\begin{aligned}
C_1 \frac{\partial^2 v_{C 1}(t)}{\partial^2 t}=
    & -\left(\frac{\partial s_{p r i}(t)}{\partial t}-\frac{s_{p r i}(t) R_L}{L}\right)     i_L(t)+\\ 
    &\frac{\partial i_1(t)}{\partial t} -\frac{s_{p r i}(t)}{L}\left(v_p(t)-n v_s(t)\right) \\
\end{aligned}
\label{algo:C1Del2}
\end{equation}

\begin{equation}
\begin{aligned}
C_2 \frac{\partial^2 v_{C 2}(t)}{\partial^2 t}= 
    & n\left(\frac{\partial s_{s e c}(t)}{\partial t}-\frac{s_{s e c}(t) R_L}{L}\right) i_L(t)-\\
    &\frac{\partial i_2(t)}{\partial t} +\frac{n s_{s e c}(t)}{L}\left(v_p(t)-n v_s(t)\right)
\end{aligned} 
\label{algo:C2Del2}
\end{equation}
where $C_1$ and $C_2$ represent input and output capacitors, and $s_{pri}(t)$ and $s_{sec}(t)$ are functions describing circuit switching behaviors. $v_{C1}$ and $v_{C2}$ govern the stability, robustness, and power quality when interfacing other power conversion circuits.

\textbf{CirNet for System-level Modeling of Circuit Physics}: Subsequently, the system-level modeling is attained with the physics-informed circuit net, CirNet. CirNet retrofits a recurrent LN-GRU net to encode circuit physical laws in its inherent feature space. Taking non-resonant DAB converter as an example, main state waveforms are $i_L$, $v_{C1}$ and $v_{C2}$. By leveraging the Kirchhoff’s, Faraday’s, and Gauss’s laws, the system-level dynamic equations are derived in Eqs.~\ref{algo:LDel} to \ref{algo:C2Del2}, where the electrical notations are given in Figure~\ref{fig:1}. Eq.~\ref{algo:LDel} describes the dynamics of high-frequency ac current $i_L(t)$, and Eq.~\ref{algo:C1Del2} and Eq.~\ref{algo:C2Del2} present the electrical behaviors of input and output capacitors, which follow second-order differential equations.

To predict the key state waveforms, CirNet takes the outputs of ModNet as its inputs and iteratively infers the next state based on the predictions of previous states, as the structure in Figure~\ref{fig:3} Part (3) and Eq.~\ref{algo:LCir} shown. 

In terms of the training for CirNet, similar to ModNet, the loss based on physics information $l_p$ and the loss based on experimental data $l_d$ are both deployed, as shown in Eq.~\ref{algo:LCir}, in which $\lambda_d$ and $\lambda_p$ are their loss factors. In $l_p$, the circuit physical dynamics expressed in differential equations~\ref{algo:LDel} to \ref{algo:C2Del2} are embedded in the loss functions for physics learning. 

In $l_d$, several data points of the inductor currents $i_L$ in the hardware experiments are taken as the ground truth, where the datapoints are denoted as $\left\{i_{L, j}^*\left(t_1\right), \ldots, i_{L, j}^*\left(t_T\right)\right\}_j^N$. The waveform predictions are saved for a switching period for the performance evaluation.  

\begin{equation}
\begin{aligned}
\hat{i}_L (t_{k+1})=\operatorname{CirNet}(
& \hat{i}_L(t_k),\operatorname{ModNet}(\\
&\hat{v}_p(t_k),\hat{v}_s(t_k) ; \theta_{{mod}}) ; \theta_{cir})
\end{aligned}
\label{algo:CirNet}
\end{equation}

\begin{equation}
\begin{aligned}
&l_{C i r} 
 =\lambda_d l_d+\lambda_p l_p \\
& =\frac{\lambda_d}{N T} \sum_j \sum_k\left[\hat{i}_{L, j}\left(t_{k+1}\right)-i_{L, j}{ }^*\left(t_{k+1}\right)\right]^2 + \\
& \frac{\lambda_p}{N T} \sum_j \sum_k\left[\begin{array}{l}
L\left(\hat{i}_{L, j}\left(t_{k+1}\right)-i_{L, j}{ }^*\left(t_k\right)\right)- \\
(\hat{v}_{p, j}\left(t_{k+1}\right)-n \hat{v}_{s, j}\left(t_{k+1}\right)- \\
R_L i_{L j}{ }^*\left(t_k\right)) \Delta t_k
\end{array}\right]^2 \\
\end{aligned}
\label{algo:LCir}
\end{equation}

\textbf{Circuit Performance Evaluation Based on CirNet}: With the output of the two hierarchical PINNs, ModNet and CirNet, the power converter performance can be evaluated. With the key waveform inference results from the previous CirNet, a variety of electrical performance metrics can be measured, aiming to provide critical and comprehensive assessments for power electronics systems. As shown in Figure~\ref{fig:3} Part (3), the difference of the maximum and minimum values of $i_L(t)$ gives the peak-to-peak current stress. The numerical integration of $[s_1(t)i_L(t)]^2$ provides the conduction loss evaluation for semiconductor switches. Soft switching performance, which is especially important for high-frequency scenarios like the electric vehicle charging, is analyzed by imposing constraints on $i_L(t)$ at the device commutation moment, where $s_{pri}(t)$ and $s_{sec}(t)$ are needed for synchronization. Other electrical performances can be gauged with the domain expertise in power electronics.  

As described in Figure~\ref{fig:2}, in the core engine, optimization algorithms are deployed to cooperate with ModNet and CirNet to find the optimal modulation parameters. The optimization procedure is an iterative cycle, in which the optimization algorithm passes the operating conditions and modulation parameters to the physics-informed surrogate models ModNet and CirNet, and these models provide evaluated performance metrics to steer the optimization process. 

By incorporating physics information within the neural networks, the model's interpretability is heightened, concurrently leading to a reduction in the necessary training data points. Furthermore, the entire design automation system exhibits seamless scalability through the integration of surrogate models for additional functions into the core engine. Importantly, users can access these functionalities without any alterations to their experience. 
\section{Experiments}
\begin{figure}[t]
    \begin{center}
        \includegraphics[width=\linewidth]{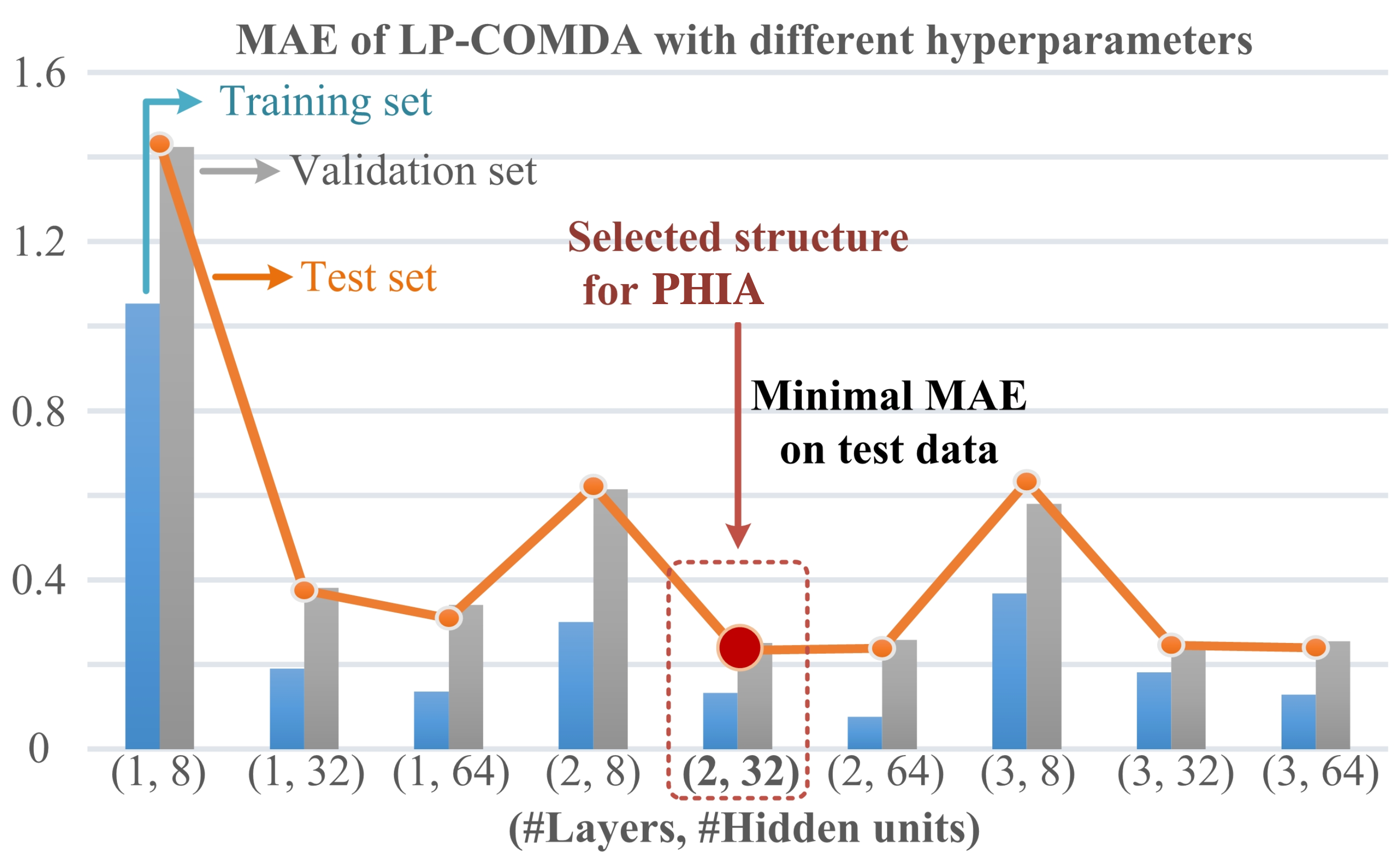}
    \end{center}
    \setlength{\abovecaptionskip}{-0.02cm}
    \caption{PHIA with different structures of CirNet. The minimal MAE of 0.235 on validation set is reached when there are 2 hidden layers with 32 hidden neurons.}
    \label{fig:3new}
\end{figure}
\begin{table}[t]
\begin{center}
\scalebox{0.93}{
\begin{tabular}{ccc}
\toprule
\textbf{Model} & \textbf{\#Parameters} & \textbf{Selected Hyperparameters} \\ \midrule
LSTM & 21,671 & 3 layers, 32 hidden size \\ \midrule
GRU & 13,319 & 1 layer, 64 hidden size \\ \midrule
LN-GRU & 13,319 & 1 layer, 64 hidden size \\ \midrule
TCN & 88,199 & \begin{tabular}[c]{@{}l@{}}2 layers, 7 kernel size, \\ 64 hidden size\end{tabular} \\ \midrule
GRU-VAE & 88,583 & 2 layers, 64 hidden size \\ \midrule
TST & 6,913 & \begin{tabular}[c]{@{}l@{}}1 layer, 2 attention heads, \\ 32 hidden size\end{tabular} \\ \midrule
TSiTPlus & 7,042 & \begin{tabular}[c]{@{}l@{}}1 layer, 2 attention heads, \\ 32 hidden size\end{tabular} \\ \midrule
MiniRocket & - & 700 features, 6 dilatation size \\ \midrule
CirNet & 9,930 & 2 layers, 32 hidden size \\ \midrule
\textbf{PHIA} & 19,857 & 2 layers, 32 hidden size \\ \bottomrule
\end{tabular}
}
\end{center}
\caption{Hyperparameter selection of PHIA and benchmarks. The chosen hyperparameters yield the lowest loss on validation set.}
\label{table:hyperparameters}
\end{table}

\subsection{Experiments Setup}

We primarily aim to ascertain the accuracy of the ModNet and CirNet surrogate models, where heightened model accuracy in the core engine translates to superior system design performance. Therefore, we carried out several experiments to validate PHIA's model performance in a low-resource setting, i.e.,  using a tiny dataset with only 200 samples.

\textbf{Data Preparation}:
In this work, we focus on a modeling task of time-series forecasting for the inductor current $i_L(t)$ of DAB converters under triple phase-shift (TPS) modulation as a representative example. Using a hardware experimental prototype, we acquired waveform data spanning 200 sequences. The configuration of the hardware prototype and the methodology employed for data acquisition are elucidated in reference \cite{10032068}.

One of the advantages of PINN models is that it requires very few samples to learn as compared to other machine learning models. To demonstrate this advantage, we conduct our experiments with two different data splitting ratios. For the first set, we split the data into 5\% training (10 samples), 10\% validation (20 samples) and 85\% test (170 samples). The second set is split into 50\%, 10\% and 40\% for training, validation and test, respectively. 

\textbf{Baseline}:
We evaluate PHIA's design outcomes using: (1) the Bayesian Network (BN) \cite{pearl1985bayesian}, (2) Support Vector Regression (SVR) \cite{smola2004tutorial}, (3) XGBoost \cite{chen2015xgboost}, (4) Random Forest (RF) \cite{breiman2001random}, (5) Long-Short-Term-Memory (LSTM) \cite{hochreiter1997long}, (6) GRU net \cite{chung2014empirical}, (7) LN-GRU net \cite{ba2016layer}, (8) Temporal Convolutional Net (TCN) \cite{lea2017temporal}, (9) GRU-Based Variational Autoencoder (GRU-VAE) \cite{an2015variational}, (10) Time-Series Transformer (TST) \cite{zerveas2021transformer}, (11) Time-Series Net Adapted from Vision Transformer (TSiT-Plus) \cite{tsai}, and (12) MiniRocket \cite{tan2022multirocket}.

\begin{figure}[t]
    \begin{center}
        \includegraphics[width=1\linewidth]{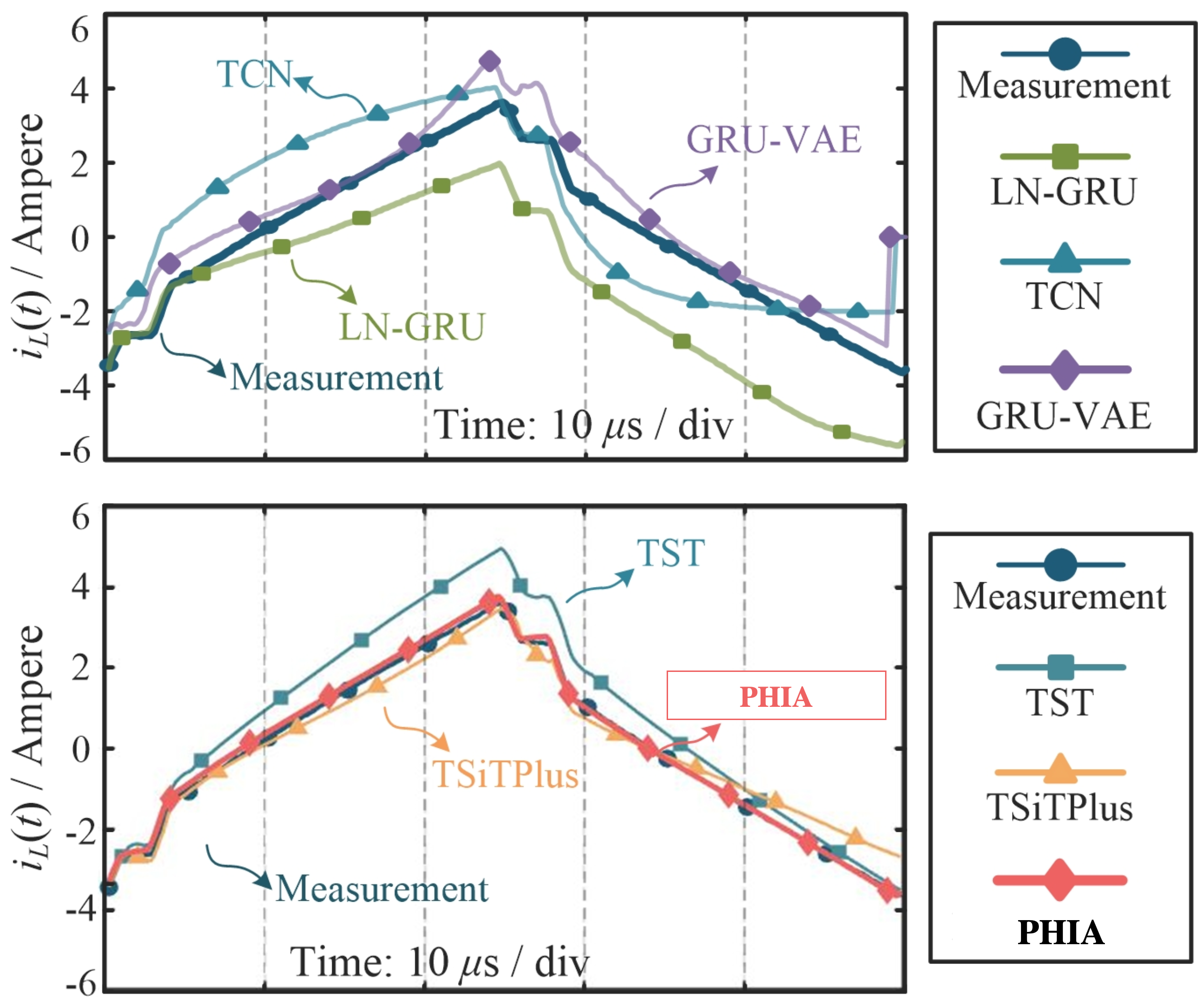}
    \end{center}
    \setlength{\abovecaptionskip}{-0.02cm}
    \caption{Modeling performance of the top 5 benchmarks and the proposed PHIA. The modelling results of PHIA is closest to the measurement, demonstrating its outstanding modeling accuracy. }
    \label{fig:11}
\end{figure}

\textbf{Hyperparameter Search}:
For fair comparison, we conduct hyperparameter search on the validation set for PHIA and all baseline algorithms. We perform grid search on important hyperparameters for each model. Figure~\ref{fig:3new} illustrates the MAEs of PHIA with different hyperparameters, shedding light on the selected best structure. The final selected sets of hyperparameters are summarized in Table~\ref{table:hyperparameters}.

\subsection{Experimental Results}
\begin{table}[t]
\begin{center}
\scalebox{0.76}{
\begin{tabular}{ccccc}
\toprule
\multirow{2}{*}{\begin{tabular}[c]{@{}c@{}}Model \end{tabular}} & \multirow{2}{*}{Training} & \multirow{2}{*}{Validation} & \multirow{2}{*}{Test} & \multicolumn{1}{l}{\multirow{2}{*}{p-value}} \\
 &  &  &  & \multicolumn{1}{l}{} \\ \midrule
BN & 2.152{$\pm$0.0} & 2.436{$\pm$0.0} & 2.515{$\pm$0.0} & 6.81E-11* \\ \midrule
SVR & 1.984{$\pm$0.0} & 2.313{$\pm$0.0} & 2.286{$\pm$0.0} & 1.16E-10* \\ \midrule
XGBoost & 2.299{$\pm$0.0} & 2.108{$\pm$0.0} & 2.792{$\pm$0.0} & 3.83E-11* \\ \midrule
RF & 4.983{$\pm$0.0} & 4.181{$\pm$0.0} & 4.434{$\pm$0.0} & 3.18E-12* \\ \midrule
LSTM & 1.164{$\pm$0.151} & 1.380{$\pm$0.042} & 1.560{$\pm$0.095} & 7.23E-8* \\ \midrule
GRU & 1.287{$\pm$0.174} & 1.336{$\pm$0.101} & 1.604{$\pm$0.064} & 6.35E-10* \\ \midrule
LN-GRU & 0.834{$\pm$0.272} & 1.289{$\pm$0.165} & 1.257{$\pm$0.117} & 1.71E-6* \\ \midrule
TCN & 1.571{$\pm$0.210} & 1.996{$\pm$0.257} & 1.538{$\pm$0.501} & 1.436E-3* \\ \midrule
TST & 0.668{$\pm$0.062} & 0.750{$\pm$0.028} & 0.666{$\pm$0.063} & 1.47E-6* \\ \midrule
TSiTPlus & 0.707{$\pm$0.158} & 1.106{$\pm$0.12} & 1.302{$\pm$0.078} & 3.46E-8* \\ \midrule
MiniRocket & 0.113{$\pm$0.122} & 0.742{$\pm$0.365} & 0.763{$\pm$0.338} & 0.0131* \\ \midrule
GRU-VAE & 1.928{$\pm$0.201} & 2.190{$\pm$0.180} & 2.219{$\pm$0.172} & 6.47E-7* \\ \midrule
\textbf{CirNet Only} & 0.121{$\pm$0.014} & 0.282{$\pm$0.014} & 0.310{$\pm$0.013} & 1.442E-3* \\ \midrule
\textbf{PHIA} & \textbf{0.140{$\pm$0.054}} & \textbf{0.235{$\pm$0.025}} & \textbf{0.245{$\pm$}0.029} & \textbf{N/A} \\ \bottomrule
\end{tabular}}
\end{center}
\caption{Experimental results with training / validation / test set split of 5\% (10 samples), 10\% (20 samples) and 85\% (170 samples), respectively. PHIA statistically outperforms baselines (marked with *), with a 63.2\% lower error as compared to the second-best benchmark (TST),measured in Mean Absolute Errors (MAE).}
\label{table:experiment}
\end{table}

\begin{table}[t]
\begin{center}
\scalebox{.76}{
\begin{tabular}{ccccc}
\toprule
\multirow{2}{*}{\begin{tabular}[c]{@{}c@{}}Model \end{tabular}} & \multirow{2}{*}{Training} & \multirow{2}{*}{Validation} & \multirow{2}{*}{Test} & \multicolumn{1}{l}{\multirow{2}{*}{p-value}} \\
 &  &  &  & \multicolumn{1}{l}{} \\ \midrule
BN & 2.455{$\pm$0.0} & 2.206{$\pm$0.0} & 2.348{$\pm$0.0} & 5.03E-14* \\ \midrule
SVR & 2.049{$\pm$0.0} & 1.780{$\pm$0.0} & 1.991{$\pm$0.0} & 1.25E-13* \\ \midrule
XGBoost & 1.321{$\pm$0.0} & 2.194{$\pm$0.0} & 1.986{$\pm$0.0} & 1.27E-13* \\ \midrule
RF & 3.167{$\pm$0.0} & 3.937{$\pm$0.0} & 3.622{$\pm$0.0} & 4.89E-15* \\ \midrule
LSTM & 1.246{$\pm$0.165} & 1.129{$\pm$0.078} & 1.193{$\pm$0.098} & 1.89E-6* \\ \midrule
GRU & 1.196{$\pm$0.127} & 1.356{$\pm$0.134} & 1.270{$\pm$0.114} & 2.80E-6* \\ \midrule
LN-GRU & 0.534{$\pm$0.097} & 0.480{$\pm$0.075} & 0.527{$\pm$0.086} & 2.33E-4* \\ \midrule
TCN & 0.685{$\pm$0.018} & 0.744{$\pm$0.062} & 0.726{$\pm$0.061} & 3.79E-6* \\ \midrule
TST & 0.264{$\pm$0.033} & 0.240{$\pm$0.033} & 0.264{$\pm$0.031} & 3.93E-3* \\ \midrule
TSiTPlus & 0.585{$\pm$0.099} & 0.567{$\pm$0.082} & 0.569{$\pm$0.099} & 2.61E-4* \\ \midrule
MiniRocket & 0.522{$\pm$0.305} & 0.616{$\pm$0.346} & 0.646{$\pm$0.343} & 0.0246* \\ \midrule
GRU-VAE & 0.686{$\pm$0.074} & 0.663{$\pm$0.036} & 0.721{$\pm$0.056} & 2.42E-6* \\ \midrule
\textbf{CirNet} Only & 0.162{$\pm$0.032} & 0.239{$\pm$0.014} & 0.234{$\pm$0.007} & 3.98E-3* \\ \midrule
\textbf{PHIA} & \textbf{0.170{$\pm$0.013}} & \textbf{0.191{$\pm$0.005}} & \textbf{0.201{$\pm$}0.006} & \textbf{N/A} \\ \bottomrule
\end{tabular}}
\end{center}
\caption{Experimental results with training / validation / test set split of 50\% (100 samples), 10\% (20 samples) and 40\% (80 samples), respectively. PHIA statistically outperforms baselines (marked with *), with a 23.7\% lower error as compared to the second-best benchmark (TST), measured in Mean Absolute Errors (MAE).}
\label{table:experiment2}
\end{table}

Table~\ref{table:experiment} and~\ref{table:experiment2} summarise the experimental results for PHIA and baseline for two different splits. Each algorithm with the respective optimal hyperparameters was ran for ten times. The average MAE for training, validation and test sets are reported. The reported P-values are the results of T-tests at a significance level of 0.02. 

For both splits, experimental results show that PHIA outperforms conventional machine learning approaches, such as BN, SVR, XGBoost and RF, and exhibits superior performance over deep learning predictive models, such as LSTM, GRU, LN-GRU and TCN, and predictive models such as GRU-VAE. When compared to state-of-the-art models like TST, TSiTPlus, and MiniRocket, PHIA again demonstrates notably enhanced forecasting accuracy. To further affirm the hierarchical PINNs structure, we contrasted CirNet with PHIA. Noticeably, PHIA with 10 training samples outperforms most benchmarks with 100 training samples, and align with CirNet-Only model. 

To visualize the modeling performance, Figure~\ref{fig:11} presents the modeled $i_L$ waveform for both PHIA and benchmark models. Notably, the PHIA modeling outcomes align more congruently with the measurements than any of the other models, demonstrating its outstanding modeling precision.

Overall, we conclude that the proposed PHIA statistically achieves better modeling results than other baseline models for power electronics modeling, even with extremely low-data scenario. 

\subsection{Interpretation}

\subsubsection{Accuracy and Reliability}
The improved accuracy observed in the experiments, particularly in low-data scenarios, underscores the robustness of the proposed framework. This enhancement ensures reliable performance, which is crucial for reducing the risk of failures in real-world applications. The system's ability to achieve high precision despite limited data sets demonstrates its adaptability and suitability for resource-constrained environments.

\subsubsection{Efficiency Gains}
The significant reduction in error rates translates directly to efficiency gains. By optimizing workflows, the framework enables engineers to achieve desired outcomes with fewer iterations and reduced computational resources. This efficiency not only accelerates the design process but also minimizes the operational costs associated with traditional methods.

\subsubsection{Scalability}
Performance improvements across metrics indicate better scalability of the framework. This scalability allows the system to handle more complex systems or broader application domains without requiring proportional increases in computational or human resources. Such scalability is essential for extending the applicability of the framework to diverse use cases.

\subsubsection{Explainability and Transparency}
The integration of hierarchical modeling enhances the explainability of the results. By aligning the model predictions with physical principles, the framework produces outputs that are both interpretable and trustworthy. This transparency is crucial for fostering confidence in the system's decisions, especially in critical domains like power electronics.

\subsubsection{Implications for Decision-Making}
The precision and reliability of the framework support informed decision-making in key applications. For example, in renewable energy systems and power grid operations, the metrics directly influence operational success and system stability. The enhanced performance ensures that the framework can be effectively deployed in high-stakes scenarios.

This interpretation highlights the broad implications of the observed performance improvements, showcasing the framework's potential to revolutionize practical applications and advance theoretical understanding in the domain.
\section{User Study}
To assess PHIA's empirical performance in improving engineering efficiency, we design a practical use case for the modulation strategy of the DAB converter and conduct an empirical experiment with 20 industrial practitioners.  

Specifically, the user requirements for this design process include: (1) Guidance in defining design specifications; (2) Optimal design outcomes for the tailored scenario; and (3) Analysis on the design outcomes. 

Based on the use case, we conduct a user study with 20 industrial practitioners recruited via research labs and collaborating external companies on a voluntary basis. The participants consist of 10 junior engineers with less than three years of experience, and 10 senior engineers with over five years of relevant experience. All participants declared to have no prior experience in using AI systems to assist engineering design works.

All participants are given two independent variants of the same task, namely to design the TPS modulation strategy for a Dual Active Bridge (DAB) converter to serve as a DC transformer of the DC-DC power grid, as shown in Fig.~\ref{fig:1}. We ask the participant to first use PHIA's chat interface to address task variant one. After completion, they then use standard practice (i.e. analytical and manual design) based on the given design requirements and data to address task variant two. To assess the effectiveness, the effective working hours are measured in 30-minute time blocks while tackling the tasks. 

Table~\ref{table:design-case} (in the Appendix) shows an example of a multi-turn conversation between the user and PHIA, as well as the output components. Empirically, the results of the user study are highly promising, in terms of both functionality and efficiency. In terms of functionality, all 20 participants are able to complete both assigned tasks successfully. In terms of efficiency, the 10 junior engineers spent an average of 1.2 time blocks and 115.5 time blocks in completing the tasks using PHIA and analytical approach, respectively; where the 10 senior engineers recorded an average of 0.9 time blocks and 30.5 time blocks with PHIA and analytical approach, respectively. Overall, we observe that Junior engineers and senior engineers experienced 96.3 times and 33.9 times efficiency, respectively.

\section{Conclusion}

Modulation strategy design of power converters is pivotal for the optimal functioning of renewable energy power systems. However, numerous AI-centric modulation design automation systems today grapple with challenges tied to explainability, usability, scalability, and data intensity. In our research, we introduced PHIA, a LLM-based Physics-informed autonomous agent to perform effective power converter modulation design automation. Our physics-informed approach combines the reasoning capabilities of large language models with rigorous physical constraints, enabling robust and interpretable design solutions. Experiments demonstrate PHIA's proficiency in accurate power converter modeling, even with a data-constrained environment. The hierarchical physics-informed neural network architecture ensures both accuracy and interpretability, while the LLM-based planning module provides an intuitive interface for engineers. Our comprehensive user study shows that using PHIA helps engineers improve efficiency by over 33 times in practice, highlighting its potential for widespread adoption in industrial applications. Future work could explore extending PHIA to other power electronics applications and incorporating additional physics-based constraints to further enhance its capabilities.
\section{Acknowledgement}

This research is supported in part by the Ministry of Education, Singapore, under its Academic Research Fund Tier 2 (Award No. MOE-T2EP20123-0015). Any opinions, findings and conclusions, or recommendations expressed in this material are those of the authors and do not reflect the views of the Ministry of Education, Singapore.

\bibliography{bibliography}

\newpage
\appendix
\begin{table*}[h]
\section{Appendix A: Example Use Case}
\vspace{1cm}
\begin{center}
\scalebox{.91}{
\begin{tabular}{cp{3cm}p{3cm}p{8cm}}
\toprule
\textbf{ID} & \textbf{Goal} & \textbf{User Input} & \textbf{LM-COMDA Output} \\ \midrule
{1} & {\begin{tabular}[c]{@{}l@{}}To define a modulation \\ strategy $S$ to design\end{tabular}} & {\begin{tabular}[c]{@{}l@{}}I need your help to design \\ my dual active bridge converter.\end{tabular}} & \begin{tabular}[c]{@{}l@{}}Sure, I can help you with that. let's start with choosing the \\ modulation strategy. Which one would you like to use for \\ your dual active bridge converter, TPS, DPS or Hybrid?\end{tabular} \\ \midrule
{2} & {\begin{tabular}[c]{@{}l@{}}To define the prioritized \\ objective $O$\end{tabular}} & {TPS.} & \begin{tabular}[c]{@{}l@{}}Great! Now let's move on to the prioritized optimization \\ objective. Which of the following objectives would you like \\ to prioritize: Efficiency,  Soft switching, or Current stress?\end{tabular} \\ \midrule
{3} & {\begin{tabular}[c]{@{}l@{}}To define operation \\ conditions\end{tabular}} & {Current stress.} & \begin{tabular}[c]{@{}l@{}}Understood.Let's discuss the operating conditions of your \\ dual active bridge converter. Could you provide the following \\ information: rated power, rated input voltage,  rated output \\ voltage, actual operating power, actual output voltage.\end{tabular} \\ \midrule
{4} & {\begin{tabular}[c]{@{}l@{}}To select optimization \\ algorithms\end{tabular}} & {\begin{tabular}[c]{@{}l@{}}Rated power: 1000 W\\ Rated input voltage: 200 V\\ Rated output voltage: 200 V\\ Operating power: 600W\\ Actual output voltage: 160V\end{tabular}} & \begin{tabular}[c]{@{}l@{}}Thank you for providing the operating conditions. Lastly, we \\ need to choose an optimization algorithm. Which optimization \\ algorithm would you choose: PSO, DE or GA?
\end{tabular} \\ \midrule
\multicolumn{1}{l}{5} & \multicolumn{1}{l}{\begin{tabular}[c]{@{}l@{}}To confirm specifications \\ are fully collected.\end{tabular}} & \multicolumn{1}{l}{PSO.} & \begin{tabular}[c]{@{}l@{}}OK, I have collected all the parameters and am about to \\ execute the program.\end{tabular} \\ \midrule
\multicolumn{4}{c}{\textbf{Design Outcome \& Analysis}} \\ \midrule
\multicolumn{4}{l}{\begin{tabular}[c]{@{}l@{}} Figure (1) presents the Current stress modeling results with respect to various modulation parameters (inner phase shifts  $D_l$ and $D_2$). \\ The optimized $D_l$ and $D_2$ for the best Current stress performance under the input voltage 200V,  power 600 W,  and the output voltage \\ 160 V is highlighted as the red dot. Figure (2) represents the current stress performance curve for the optimal TPS and the compared \\ SPS when power is within {[}100 W, 1000 W{]} range. Figure (3) is the curve of optimal modulation parameter selection for the best \\ current stress within {[}100 W, 1000 W{]} range. You can deploy the values in your controllers.
\end{tabular}} \\ \midrule
\multicolumn{2}{c|}{\includegraphics[height=42mm]{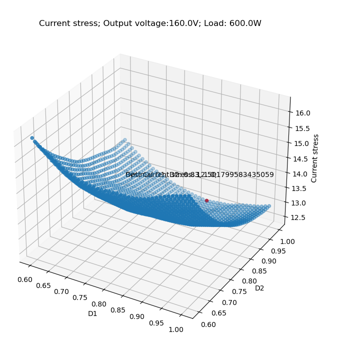}} & \multicolumn{1}{c|}{\includegraphics[height=37mm]{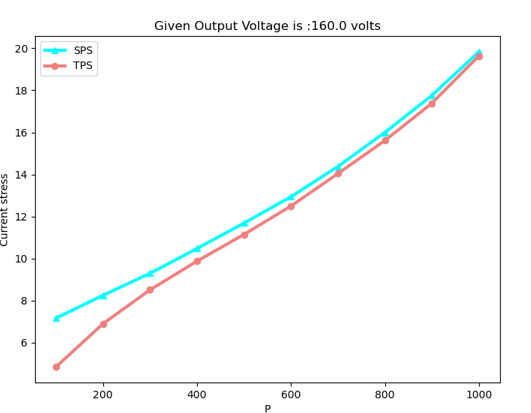}} & \multicolumn{1}{c}
{\includegraphics[height=40mm]{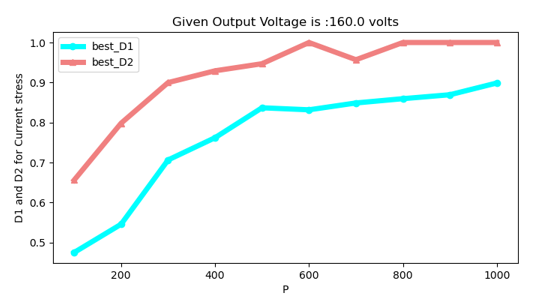}} \\
\multicolumn{2}{c|}{Figure (1)} & \multicolumn{1}{c|}{Figure (2)} & \multicolumn{1}{c}{Figure (3)} \\ \bottomrule
\end{tabular}}
\end{center}
\caption{An use case exemplar: with 5 request series, PHIA collects all necessary design specifications, which include a modulation strategy $S$ to design, prioritized objective $O$, operating conditions, and the selected optimization algorithms. After specification collection via text input, the design outcome is presented with text and figures.}
\label{table:design-case}
\end{table*}

\end{document}